\documentclass[runningheads]{llncs}
\usepackage{graphicx}
\usepackage{subfigure}

\usepackage{tikz}
\usepackage{comment}
\usepackage{amsmath,amssymb} 
\usepackage{color}
\usepackage{wrapfig}
\usepackage[accsupp]{axessibility}  
\makeatletter
\newcommand{\printfnsymbol}[1]{%
\textsuperscript{\@fnsymbol{#1}}%
}

\usepackage{graphicx}
\usepackage{amsmath}
\usepackage{amssymb}
\usepackage{booktabs}
\usepackage{algorithm}
\usepackage{algorithmic}
\usepackage{xcolor}

\begin{document}
\pagestyle{headings}
\mainmatter
\def\ECCVSubNumber{2293}  

\title{Meta Spatio-Temporal Debiasing for Video Scene Graph Generation} 

\titlerunning{Meta Spatio-Temporal Debiasing for Video Scene Graph Generation}
%
\author{Li Xu\inst{1}\thanks{Both authors contributed equally to the work.}
\and
Haoxuan Qu\inst{1}\printfnsymbol{1}
\and
Jason Kuen\inst{2}
\and
Jiuxiang Gu\inst{2}
\and
Jun Liu\inst{1}\thanks{Corresponding Author}
}
\authorrunning{L. Xu et al.}
%
\institute{Singapore University of Technology and Design \\
\email{\{li\_xu, haoxuan\_qu\}@mymail.sutd.edu.sg, jun\_liu@sutd.edu.sg} \and
Adobe Research\\
\email{\{kuen, jigu\}@adobe.com}}
\maketitle

\begin{abstract}
Video scene graph generation (VidSGG) aims to parse the video content into scene graphs,
which involves modeling the spatio-temporal contextual information in the video.
However, due to the long-tailed training data in datasets, the generalization performance of existing VidSGG models can be affected by the \textbf{spatio-temporal conditional bias} problem. 
In this work, from the perspective of meta-learning, we propose a novel Meta Video Scene Graph Generation (\textbf{MVSGG}) framework to address such a bias problem. 
Specifically, to handle various types of spatio-temporal conditional biases, our framework first constructs a support set and a group of query sets from the training data, where the data distribution of each query set is different from that of the support set w.r.t. a type of conditional bias. 
Then, by performing a novel \textbf{meta training and testing} process to optimize the model to obtain good testing performance on these query sets after training on the support set, our framework can effectively guide the model to learn to well generalize against biases.
Extensive experiments demonstrate the efficacy of our proposed framework.
\end{abstract}

\section{Introduction}\label{sec:intro}
A scene graph is a graph-based representation, which encodes different visual entities as nodes and the pairwise relationships between them as edges, i.e., in the form of \texttt{subject predicate object} relation triplets \cite{johnson2015image,shang2017video}. 
Correspondingly, the task of video scene graph generation (VidSGG) aims to parse the video content into a sequence of spatio-temporal relationships between different objects of interest \cite{xu2017scene,shang2017video}.
Since it can provide refined and structured scene understanding, the video scene graph representation has been widely used in
various higher-level video tasks, 
such as video question answering \cite{lei2018tvqa,tapaswi2016movieqa}, video captioning \cite{guadarrama2013youtube2text,venugopalan2015sequence}, and video retrieval \cite{lei2020tvr,gao2017tall}. 

However, despite of the great progress of VidSGG \cite{teng2021target,cong2021spatial,chen2021social,lu2021context},
most existing approaches tackling this task may suffer from the problem of \textit{spatio-temporal conditional biases}.
Specifically, as shown by previous works \cite{tang2020unbiased,chiou2021recovering,li2021interventional},
there exist long-tailed training data issues in existing SGG datasets.
While in the context of VidSGG, given the complex spatio-temporal nature of this task, such long-tailed issues can lead to spatio-temporal conditional biases that affect the model generalization performance.
Here conditional biases mean the problem that once the model detects  certain context information (i.e., conditions) in the visual content, it is likely to directly predict certain labels (i.e., biased prediction), which however may contradict with the ground-truth.
Some works \cite{seo2022information,ye2021adversarial} also refer to this problem as spurious correlation.
In particular, in the VidSGG task, this conditional bias issue can be further divided into two sub-problems: temporal conditional bias and spatial conditional bias.

\setlength\intextsep{0pt}
\begin{wrapfigure}[27]{r}{0.5\textwidth}
  \centering
  \subfigure[Example of temporal conditional bias.]{\label{fig:short-b}\includegraphics[width=0.5\textwidth]{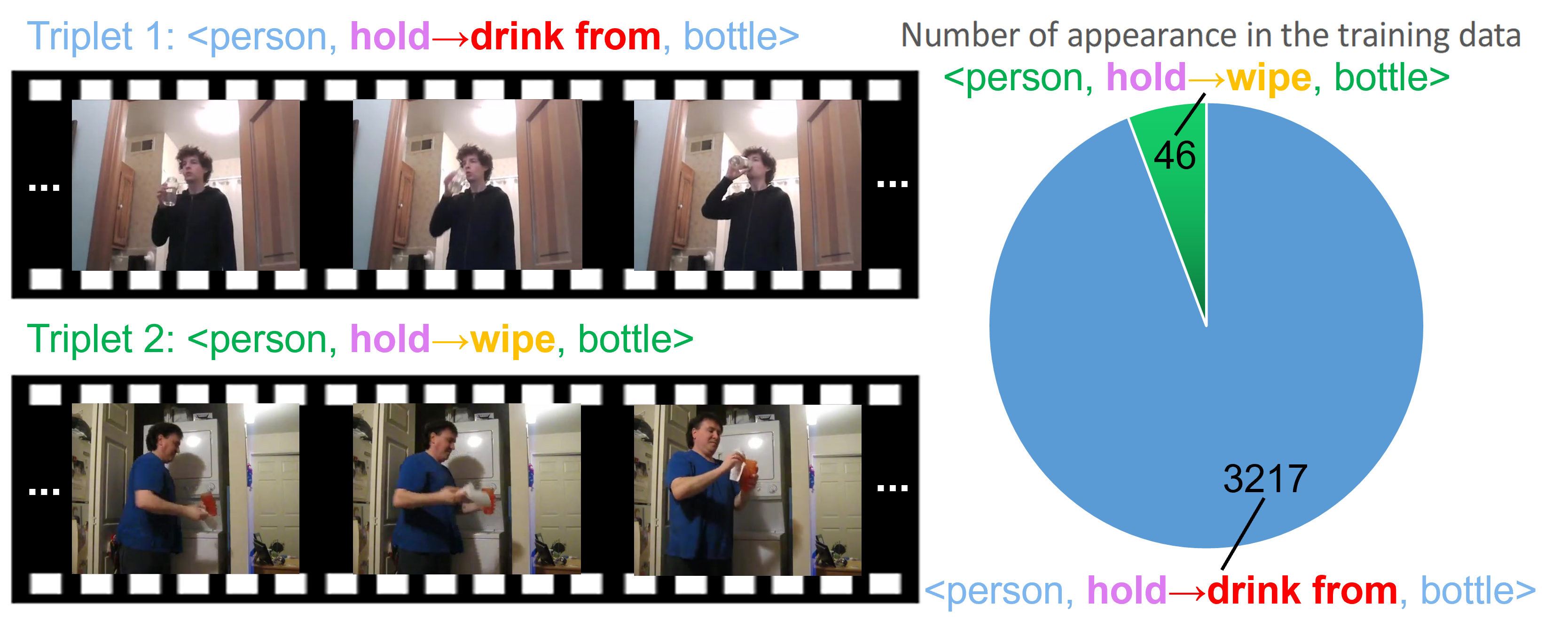}}
  \subfigure[Example of spatial conditional bias.]{\label{fig:short-a}\includegraphics[width=0.5\textwidth]{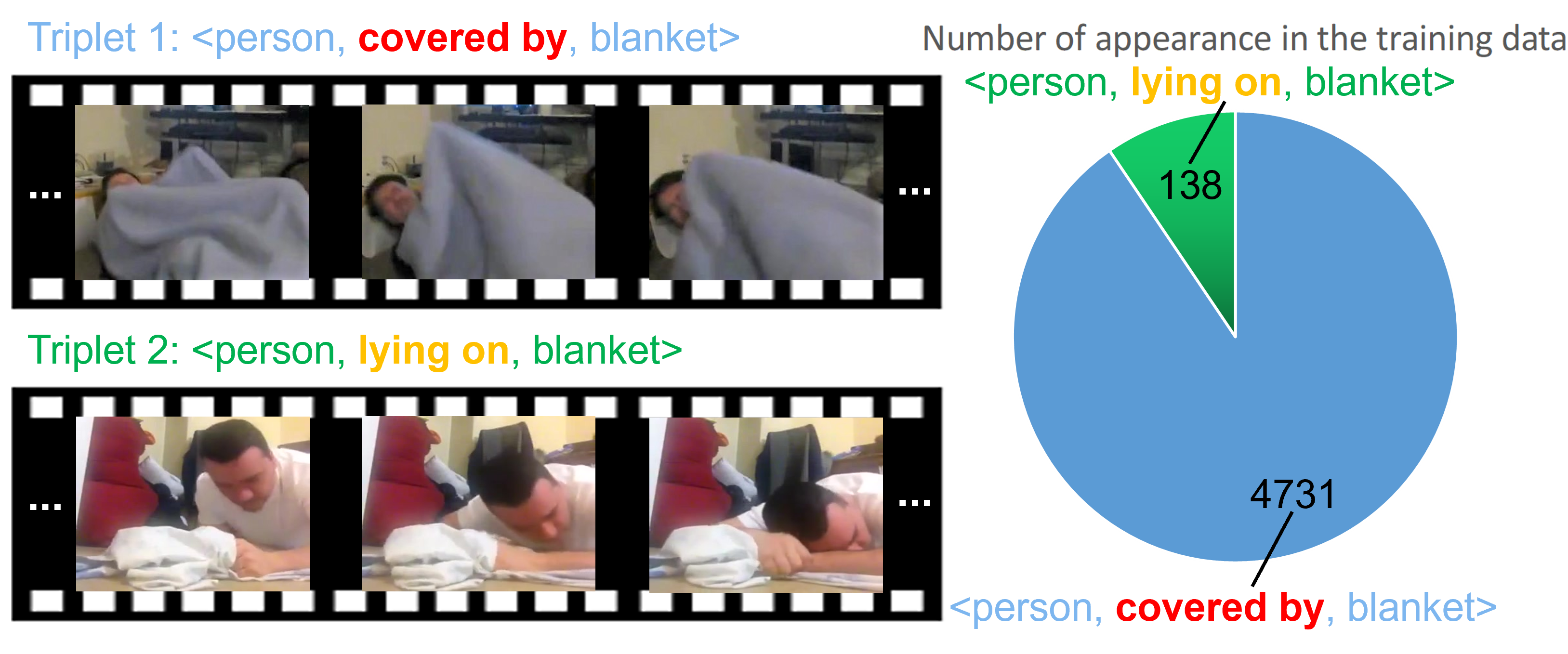}}
   \caption{\textbf{Illustration of spatio-temporal conditional biases examples from Action Genome dataset \cite{ji2020action}}.
   (a) illustrates an example of temporal conditional bias towards \texttt{person drink from bottle} conditioned on \texttt{person hold bottle} in consecutive video parts, and
   (b) presents an example of spatial conditional bias towards the predicate of \texttt{covered by} conditioned on the subject-object pair of \texttt{person} and \texttt{blanket}.
   Such spatio-temporal conditional biases can affect the performance of VidSGG models.} 
   \label{fig:example}
\end{wrapfigure}

For \textbf{temporal} conditional bias, as shown in the example of Fig. \ref{fig:example} (a),
if \texttt{person hold bottle} appears first (i.e., \textbf{temporal} context) in the video,
and in most cases, \texttt{person drink from bottle} happens next,
then a conditional bias can be established towards \texttt{person drink from bottle} conditioned on \texttt{person hold bottle} 
along the temporal axis in the video.
Due to such temporal conditional bias, 
once the previous video part involves \texttt{person hold bottle}, 
the trained model is very likely to simply predict \texttt{person drink from bottle} for the next part, which however can contradict with the ground truth as in the example.

Such a conditional bias problem also exists in the \textbf{spatial} domain when tackling the VidSGG task.
For example, as shown in Fig. \ref{fig:example} (b),
there are many more videos containing \texttt{person covered by blanket} than \texttt{person lying on blanket} in the VidSGG dataset: Action Genome \cite{ji2020action}. 
This can lead to a conditional bias
towards the predicate (\texttt{covered by}) given the \textbf{spatial} contexts of \texttt{person} and \texttt{blanket}.
Due to such spatial conditional bias,
once \texttt{person} and \texttt{blanket} appear in the video, 
the model tends to directly predict \texttt{person covered by blanket} that however can be incorrect. 

We observe that such spatio-temporal conditional bias problems widely exist in VidSGG datasets
\cite{shang2017video,ji2020action}.
Meanwhile, to correctly infer the relation triplets,
VidSGG models need to effectively model the spatio-temporal context information in the video \cite{teng2021target,cong2021spatial,shang2017video}. 
Thus the models can be easily prone to exploiting the conditional biases w.r.t. 
the relation triplet components (e.g., the predicate) 
based on the spatio-temporal contexts during training, and then fail to generalize to
data samples in which 
such conditional biases no longer hold as shown in Fig. \ref{fig:example}. 
Therefore, to address this issue for obtaining better generalization performance, we propose a novel meta-learning based framework, Meta Video Scene Graph Generation (\textbf{MVSGG}).

Meta learning, also known as learning to learn, aims to enhance the model generalization capacity
by incorporating \textit{virtual testing} during model training \cite{finn2017model,nichol2018first,huang2021metasets}.
Inspired by this, 
to improve the generalization performance of VidSGG models against the conditional biases, 
our framework incorporates a \textit{meta training and testing} scheme.
More concretely, 
we can split the training set to construct a support set for \textit{meta training} and a query set for \textit{meta testing}, 
which have different data distributions w.r.t. the conditional biases, i.e., creating a virtual testing scenario where 
simply exploiting the conditional biases during training would lead to poor testing performance.
For example, given the same subject-object pair of \texttt{person} and \texttt{blanket},
if the support set contains more \texttt{person covered by blanket} relation triplets, 
the query set 
can contain more \texttt{person lying on blanket} triplets on the contrary. 
We first use the support set to train the model (i.e., \textit{meta training}), and then evaluate the trained model on the query set (i.e., \textit{meta testing}).
According to the evaluation performance (loss) on the query set, we can further update the model 
to obtain better generalization performance. 
Since the query set is distributionally different from the support set w.r.t. the conditional biases,
by improving the testing performance on the query set after training on support set 
via \textit{meta training and testing},
our model is driven to learn to capture the ``truly" generalizable features in the data instead of relying on the biases. 
Thus our model can ``learn to generalize'' well, 
even when handling the ``difficult'' testing samples that contradict the biases in training data.

Moreover, there can exist various types of conditional biases besides the ones shown in Fig. \ref{fig:example}.
For example, there can also exist a conditional bias w.r.t. the object based on subject-predicate pair in relation triplets,
if an object appears more frequently given the same subject-predicate pair in the training data.
Thus to better handle such a range of conditional biases,
we can construct a group of query sets, where each query set is distributionally different from the support set w.r.t. one type of conditional bias.
In this manner, by utilizing all these query sets to improve the model generalization performance via \textit{meta training and testing},
our framework can effectively address various types of conditional biases in the video, and enhance the robustness of the VidSGG model.

Our MVSGG framework is general since it only changes the model training scheme (i.e., via \textit{meta training and testing}),
and thus can be flexibly applied to various off-the-shelf VidSGG models. 
We experiment with multiple models, and achieve consistent improvement of model performance.

The contributions of our work are summarized as follows.
1) 
We propose a novel \textit{meta training and testing} framework, MVSGG, for effectively addressing the spatio-temporal conditional bias problem in VidSGG.
2) By constructing a support set and multiple query sets w.r.t. various types of conditional biases, 
our framework can enable the trained model to learn to generalize well against various types of conditional biases simultaneously. 
3) Our framework achieves significant performance improvement when applied on state-of-the-art models on the evaluation benchmarks \cite{shang2017video,ji2020action}.

\section{Related Work}\label{sec:related_work}
\textbf{Scene Graph Generation (SGG).} 
Being able to provide structured graph-based representation of an image or a video, 
scene graph generation (SGG) has attracted extensive research attention \cite{johnson2015image,tang2019learning,xu2017scene,li2021bipartite,suhail2021energy,guo2021general,chen2019knowledge,liu2021fully,shang2017video,teng2021target}.
For image SGG (ImgSGG), a variety of methods \cite{tang2019learning,zhang2019graphical,zellers2018neural,yu2020cogtree} have been proposed.
Suhail et al. \cite{suhail2021energy} proposed an energy-based framework to improve the model performance by learning the scene graph structure.
Yang et al. \cite{yang2021probabilistic} investigated the diverse predictions for predicates in SGG from a probabilistic view.

Besides ImgSGG, there are also increasing research efforts exploring the task of video scene graph generation (VidSGG) \cite{shang2017video,liu2020beyond,chen2021social,teng2021target}.
This task provides two task settings based on the granularity of the generated video scene graphs: video-level \cite{shang2017video,tsai2019video,qian2019video,shang2021video,liu2020beyond,chen2021social} and frame-level \cite{teng2021target,cong2021spatial}.
For video-level VidSGG, models generate scene graphs based on the video clip, where each node encodes the spatio-temporal trajectory of an object, and the connecting edge denotes the relation between two objects. 
Shang et al. \cite{shang2017video} first investigated this problem setting, and proposed to extract improved Dense Trajectories features \cite{wang2013action} for handling this problem. 
Later on, some other methods have been proposed to solve this video-level VidSGG problem from different perspectives, including the fully-connected spatio-temporal graph \cite{tsai2019video}, 
and iterative relation inference \cite{shang2021video}.
For frame-level VidSGG, a scene graph is generated for each video frame \cite{teng2021target,cong2021spatial}. 
To handle this problem setting, Teng et al. \cite{teng2021target} proposed to use a hierarchical relation tree to capture the spatio-temporal context information. 
Cong et al. \cite{cong2021spatial}  proposed to solve this problem via a spatio-temporal transformer.

For SGG, there often exists the long-tailed data bias issue that hinders models from obtaining better performance.
To solve this problem, various debiasing methods have been proposed.
Tang et al. \cite{tang2020unbiased} introduced a debiasing framework by utilizing the Total Direct Effect (TDE) analysis.
Guo et al. \cite{guo2021general} proposed a balance adjustment method to handle this issue.
Li et al. \cite{li2021interventional} explored a causality-inspired interventional approach to reduce the data bias in VidSGG.
Differently, to cope with the spatio-temporal conditional bias problem in SGG, from the perspective of meta learning, we propose a novel learning framework that can train the SGG model to learn to better generalize against biases.

\textbf{Meta Learning.}
As a group of representative works in meta learning, 
MAML \cite{finn2017model}
and its following works \cite{nichol2018first,sun2019meta,rajeswaran2019meta} mainly tackle the few-shot learning problem.
These approaches often need to perform test-time model update for fast adaptation to new few-shot tasks.
While recently, meta learning techniques have also been explored in other tasks \cite{bai2021person30k,Li_Yang_Song_Hospedales_2018,huang2021metasets,guo2020learning,qu2022improving} to improve the model performance without the need of test-time update, such as in domain generalization \cite{bai2021person30k} and point cloud classification \cite{huang2021metasets}. 
Different from existing works, here to address the challenging spatio-temporal conditional bias problem in SGG, we propose a framework that optimizes the model via \textit{meta training and testing} over the constructed support set and query sets with different data distributions w.r.t. the conditional biases.

\begin{figure*}[t]
  \centering
  \includegraphics[width=1.0\textwidth]{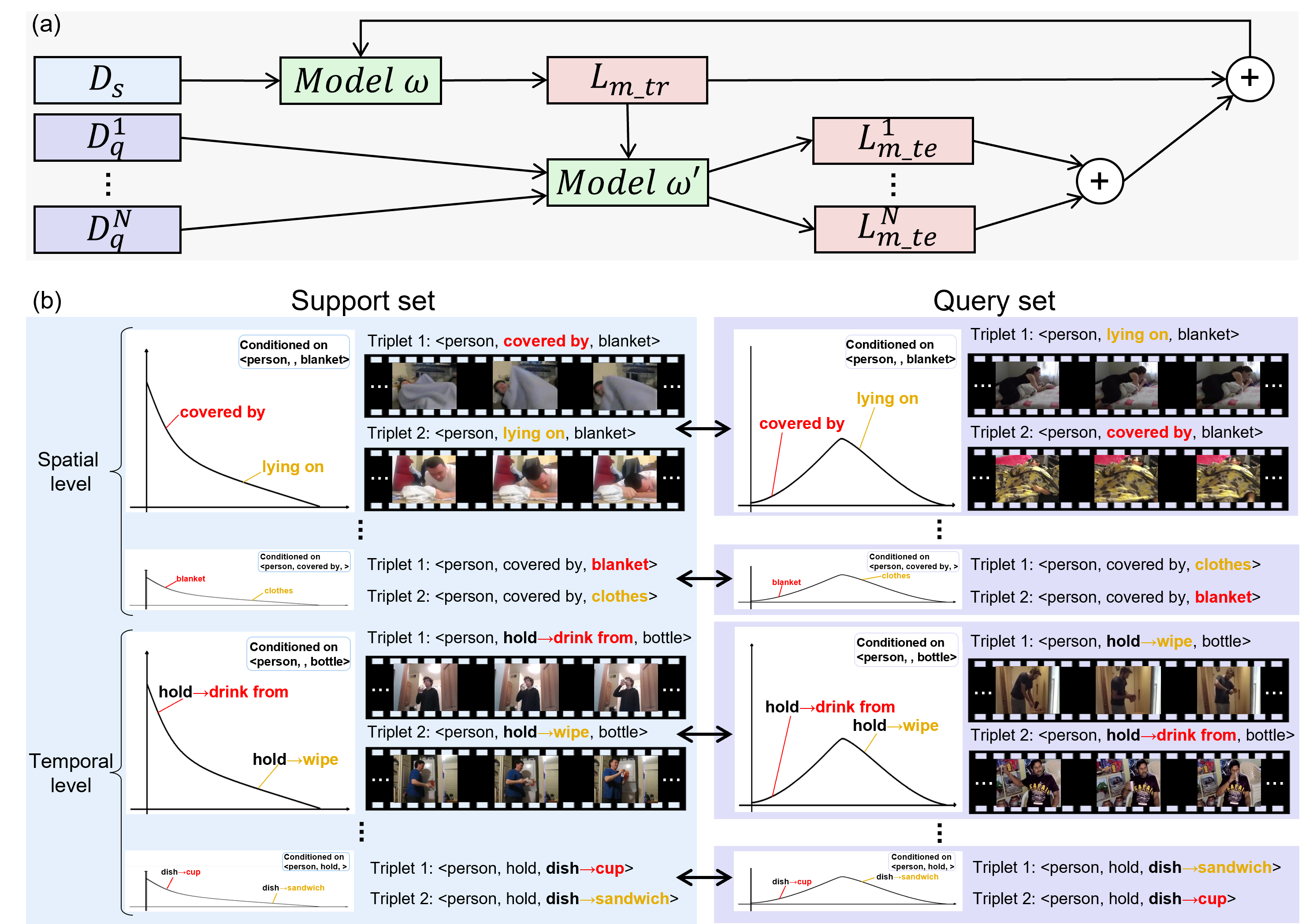}
   \caption{\textbf{Overview of our framework.}
    (a) illustrates our \textit{meta training and testing} scheme. 
    1) We first train the model (with parameters $\omega$) on the support set ($D_s$) by optimizing the loss function ($L_{m\_tr}$), and thus obtain the model with updated parameters ($\omega’$), i.e., \textit{meta training} process. 
    2) We evaluate the updated model on a group of query sets ($\{{D^{n}_q}\}^N_{n=1}$), by computing the losses ($\{{L_{m\_te}^n(\omega’)}\}_{n=1}^N$) on these query sets (i.e., \textit{meta testing} process).
    3) Finally, based on the evaluation losses, we perform meta-optimization to update the model to improve its generalization performance.
    (b) shows that to address various types of \textbf{spatial level} and \textbf{temporal level} conditional biases, 
    we construct a support set and a group of query sets for \textit{meta training and testing}. 
    The data distribution of each query set (see right side of (b)) is different from that of the support set (see left side of (b)) w.r.t. a type of conditional bias.
    }
   \label{fig:meta}
\end{figure*}
\section{Method}
As discussed above, affected by the conditional biases in the dataset,
the VidSGG model can fail to generalize to the data samples where the biases do not hold.
To address this problem, we aim to train the model to learn generalizable features in the data instead of exploiting biases.
Here generalizable features refer to the learned features that can enable the model to make unbiased predictions, i.e., obtaining robust performance. 
How to achieve this goal?
We notice that some meta-learning works \cite{finn2017model,nichol2018first,sun2019meta}
propose to boost the model learning ability via \textit{meta training and testing}.
Concretely, these works use \textit{meta training and testing} to mimic the model training and testing for improving model generalization performance.
Inspired by this, 
we propose a novel MVSGG framework, that optimizes the VidSGG model via \textit{meta training and testing} for robust generalization performance against the biases.

More specifically,
our framework first splits the training set ($D_{train}$) into a support set ($D_{s}$) for \textit{meta training}, and a group of ($N$) query sets ($\{{D^{n}_q}\}^N_{n=1}$) for \textit{meta testing},
where each query set is distributionally different from the support set w.r.t. one type of conditional bias. 
Then we first train the model using the support set (i.e., \textit{meta training}), and then evaluate the model testing performance on each of the query sets (i.e., \textit{meta testing}).
Since the biases in meta-training data (support set) do not hold in meta-testing data (query sets) due to their different data distributions, 
if the model trained on the support set can still obtain good testing performance under this condition, 
it indicates the model has learned more generalizable features rather than biases during the training process.
As a result, we can optimize the model performance in \textit{meta testing}, which can serve as a \textit{generalization feedback}, to drive and adjust the model training on the support set towards learning more generalizable features.
Below, we first introduce the \textit{meta training and testing} scheme in our framework, and then describe how to construct the support set and query sets.

\subsection{Meta Training and Testing}
{\bf Meta training.}
Using the support set $D_s$, we first 
train a VidSGG model (with parameters $\omega$), via conventional gradient update.
Specifically, we 
compute the model loss on the support set as:
\begin{equation}\label{eq:str_loss}
\setlength{\abovedisplayskip}{3pt}
\setlength{\belowdisplayskip}{3pt}
    L_{m\_tr}(\omega)= L(D_s; \omega)
\end{equation}
where $L(\cdot)$ denotes the loss function (e.g., cross-entropy loss) for training the VidSGG model.
Then we update the model parameters via gradient descent as: 
\begin{equation}\label{eq:str_update}
\setlength{\abovedisplayskip}{3pt}
\setlength{\belowdisplayskip}{3pt}
\omega’ = \omega - \alpha\nabla_\omega L_{m\_tr}(\omega)
\end{equation}
where $\alpha$ is the learning rate for \textit{meta training}. 
Note that the parameters update in this step is virtual (i.e., the updated parameters $\omega’$ is merely intermediate parameters), and the actual update for parameters $\omega$ will be performed in the meta-optimization step. 

{\bf Meta testing.} 
After \textit{meta training} on the support set ($D_s$), we then evaluate the generalization performance of the model with the updated parameters ($\omega’$), on the query sets ($\{{D^{n}_q}\}^N_{n=1}$).
In particular, for each query set $D^{n}_q$, we 
compute the model loss $L_{m\_te}^n$ on this query set as:
\begin{equation}\label{eq:meta-test}
\setlength{\abovedisplayskip}{3pt}
\setlength{\belowdisplayskip}{3pt}
  L_{m\_te}^n(\omega’)= L(D^n_q; \omega’)
\end{equation}
This computed loss can measure the model generalization performance on the query set after training on the support set, and will be used to provide feedback on \textit{how the model should be updated so that it can generalize to different data distributions against the biases} in the following meta-optimization step. 

{\bf Meta-optimization.} 
As discussed above, we aim to optimize the model parameters ($\omega$), so that \textit{after the training (update) on the support set} (\textit{i.e.,} $\omega\rightarrow\omega’$), \textit{it can also obtain good testing performance} (\textit{i.e., lower $L_{m\_te}^n(\omega’)$}) \textit{on all the query sets against the biases in training data}.
Towards this goal, inspired by MAML \cite{finn2017model}, the meta-optimization objective can be formulated as:
\begin{equation}\label{eq:meta-optimization}
\setlength{\abovedisplayskip}{6pt}
\setlength{\belowdisplayskip}{6pt}
\begin{aligned}
    &\min_{\omega}\;L_{m\_tr}(\omega) + \sum_{n=1}^N L_{m\_te}^n(\omega’) \\
    =&\min_{\omega}\;L_{m\_tr}(\omega) + \sum_{n=1}^N L_{m\_te}^n \big (\omega - \alpha\nabla_\omega L_{m\_tr}(\omega) \big)
\end{aligned}
\end{equation}
where the first term denotes the model training performance,
while the second term denotes the model \textit{generalization} performance (with the updated parameters $\omega’$).
Note that the above meta-optimization is performed over the initial model parameters $\omega$, while $\omega’$ is merely intermediate parameters for evaluating the model generalization performance ($L_{m\_te}^n(\omega’)$) during \textit{meta testing}.
Based on the meta-optimization objective in Eqn. \ref{eq:meta-optimization}, we can update the model parameters $\omega$ as:
\begin{equation}\label{eq:meta-update}
\setlength{\abovedisplayskip}{5pt}
\setlength{\belowdisplayskip}{5pt}
\begin{aligned}
    \omega \leftarrow \omega - \beta\nabla_\omega \Big (L_{m\_tr}(\omega) + \sum_{n=1}^N L_{m\_te}^n \big (\omega - \alpha\nabla_\omega L_{m\_tr} (\omega) \big) \Big)
\end{aligned}
\end{equation}
where $\beta$ denotes the learning rate for meta-optimization. 
Via such optimization, the model is driven to learn to capture more generalizable features to generalize well against biases.

Here we provide an intuitive analysis of such meta-optimization.
During the above ``learning to learn'' process, the model is first trained (updated) over the support set (i.e., $\omega\rightarrow\omega’$).
In this step, 
the biases in meta-training data (support set) can be learned by the model, since such biases can contribute to model performance on meta-training data.
However, 
to generalize well to the meta-testing data (query sets) where biases in meta-training data (support set) no longer hold, 
the model needs to learn to avoid learning biases and instead capture more generalizable features during \textit{meta training}.
This means that the second term in Eqn. \ref{eq:meta-update} that involves the second-order gradients of $\omega$ (i.e., meta-gradients): $\nabla_\omega L_{m\_te}^n \big (\omega - \alpha\nabla_\omega L_{m\_tr}(\omega) \big )$, serves as a \textit{generalization feedback} 
to the model learning process ($\omega\rightarrow\omega’$) on the support set about 
how to learn more generalizable features. 

From the above analysis, we can also conclude that the efficacy of our framework for debiasing lies in the simulated difficult testing scenarios where training data biases no longer hold.
This  also implies that we are not using the query sets to simulate the data distribution of the real testing set, which is unknown during model training.
Instead, we only need to make the difference between the data distribution of meta-testing data (query sets) and that of meta-training data (support set) to be as large as possible w.r.t. the biases, so as to drive the model learning to learn more generalizable features.
We also provide theoretical analysis of the efficacy of this framework for alleviating the bias learning in the supplementary.
We perform the above three steps (i.e., meta training, meta testing and meta-optimization) iteratively until the model converges.

\subsection{Dataset Split}\label{sec:dataset_split}
As mentioned above, to handle various types of conditional biases, we split the original training set to construct a support set and a group of $N$ query sets for \textit{meta training and testing}.
In this way, the purpose of the following dataset split strategy is to make each query set distributionally different from the support set w.r.t. one type of conditional bias.
Under the guidance of this strategy, we can easily construct the support set and query sets.
Below we first discuss the details of the support set and query sets, and then introduce the strategy for constructing each query set by selecting the data samples, of which the data distributions have the largest KL divergences to the support set w.r.t. the corresponding type of conditional bias.
Some visualization examples of data distributions of the support set and query sets can refer to supplementary.

{\noindent\bf Support Set and Query Sets.} 
We first randomly select a part of the training set data as the support set ($D_{s}$), and the remaining part of the training set will be used to construct various query sets ($\{{D^n_q}\}_{n=1}^N$), where each query set is designed to address one type of conditional bias.
Since the conditional biases in VidSGG can be roughly grouped into the spatial level and the temporal level, we correspondingly construct our query sets based on these two levels, as follows.

\textbf{Spatial level.}
There can exist conditional biases between a part of the relation triplet (e.g., the predicate) and the remaining parts (i.e., the spatial contexts).
For example, as shown in Fig. \ref{fig:example} (b), given the same subject-object pair of \texttt{human} and \texttt{blanket}, the corresponding predicate is \texttt{covered by} in most triplets.
To reduce such spatial conditional bias w.r.t. the \textbf{\underline{predicate} conditioned on subject-object pair}, we can construct a query set, in which the distribution of the predicates conditioned on the same subject-object pair is different from the support set, as shown in Fig. \ref{fig:meta}.

Similarly, conditional bias can also exist w.r.t. the \textbf{\underline{predicate} conditioned on subject}.
For instance, if there are many more triplets containing \texttt{bear play} than the triplets containing \texttt{bear bite} in the dataset,
a conditional bias can be established towards the predicate \texttt{play} given the subject \texttt{bear}.
Thus we can build a query set where the distribution of the predicates (e.g., \texttt{play}, \texttt{bite}) conditioned on the same subject (e.g., \texttt{bear}) is different from that of the support set.
In a similar way, we can also construct a query set to handle the conditional bias w.r.t. the \textbf{\underline{predicate} conditioned on object}. 

Therefore, 3 query sets can be constructed to handle the corresponding 3 types of conditional biases w.r.t. the \textit{predicate} (\textit{predicate-centered}) conditioned on other parts of the relation triplet (i.e., the subject-object pair, or the subject, or the object), as discussed above.
Similarly, when considering the conditional biases w.r.t. the \textit{subject} (\textit{subject-centered}) conditioned on other parts of the triplet, 
we can also construct 3 query sets, and the same goes for the \textit{object-centered} scenario.
Thus we will construct a total of \textbf{9} query sets for handling these different types of spatial conditional biases.

\textbf{Temporal level.}
In VidSGG, 
when predicting a relation triplet, besides spatial contexts, there can also exist conditional biases between the current triplet and its temporal contexts.
Specifically, temporal conditional bias can exist between the current triplet and the triplets that appear before it,
and for simplicity, we refer to this case as \textit{forward case}.
Similarly, conditional bias can also exist between the current triplet and the triplets that appear after it (\textit{backward case}).
For these two cases, the query set construction procedures are similar,
and below we take the forward case as the example to describe such procedures.

For example, as shown in Fig. \ref{fig:example} (a), if \texttt{human hold bottle} happens first in the video,
and then \texttt{human drink from bottle} follows in most cases,
then there can exist temporal conditional bias between the \textbf{previous predicate} (e.g., \texttt{hold}) \textbf{and current predicate} (e.g., \texttt{drink from}), based on the subject-object pair (e.g., \texttt{human} and \texttt{bottle}).
To handle such temporal conditional bias,
we can construct a query set, in which the distribution w.r.t. the temporal change of predicates, 
conditioned on the same subject-object pair,
is different from that in the support set.
For example, if the support set has more videos containing \texttt{human hold bottle}$\rightarrow$\texttt{human drink from bottle}, the query set will involve more videos containing other cases w.r.t. the temporal change of predicates, such as \texttt{human hold bottle}$\rightarrow$\texttt{human wipe bottle}.

Similarly, temporal conditional bias can also exist between the \textbf{previous subject and current subject},
and we can construct a query set for handling this type of conditional bias.
In a similar manner, we can also construct a query set for handling the temporal conditional bias between the \textbf{previous object and current object}. 
Therefore, we construct 3 query sets to handle the above 3 types of temporal conditional biases in the \textit{forward case}. 
Similarly, we can also construct 3 query sets for the \textit{backward case}, and thus a total of \textbf{6} query sets for handling various types of temporal conditional biases can be obtained.

As a result, considering both spatial-level and temporal-level conditional biases, 
we construct \textbf{9+6=15} query sets ($N$=15) in total from the training data.

{\noindent\bf Query Sets Construction Strategy.} 
For constructing each query set, we need to select suitable video samples from the candidate video samples, 
so that the difference between the data distribution of the triplets in the selected videos (i.e., query sets) and that of the support set is as large as possible w.r.t. the corresponding type of conditional bias.
Here to achieve this goal, 
we adopt an efficient and generalizable strategy by maximizing the KL divergence between the data distributions of the query set and support set w.r.t. the biases, which can be applied to construct each of the 15 query sets.
Below we take the process of constructing the query set for handling the spatial conditional biases w.r.t. the \textbf{predicate conditioned on subject} as an example, to describe such a strategy.

As mentioned before, for handling this conditional bias,
we aim to construct a query set, of which the distribution of the \textbf{predicates} (e.g., \texttt{play}, \texttt{bite}) given the same \textbf{subject} (e.g., \texttt{bear}), is different from the support set.
For simplicity, we use $\phi_q$ to denote such a distribution of the query set, and $\phi_s$ to denote this distribution of the support set.
Then since the KL divergence can be used to measure the difference of two distributions, we here aim  
to construct a query set, so that the KL divergence between $\phi_q$ and $\phi_s$ (i.e., $D_{KL}(\phi_q \parallel \phi_s)$) is large.

To this end, we perform the following four steps.
(1) We first compute the distribution $\phi_s$, 
i.e., computing the probability of the occurrence of each \textbf{predicate} conditioned on the same \textbf{subject} (e.g., $p(\texttt{play}|\texttt{bear}$), $p(\texttt{bite}|\texttt{bear}$)) in the support set. 
(2) Then, assuming we have a total of $N_c$ candidate video samples for constructing the query sets,
since each candidate video sample ($i$) contains multiple relation triplets,
we can also compute its corresponding data distribution ($\phi_c^i$, $i\in\{{1,...,N_c}\}$).
(3) Since we aim to select a set of video samples to construct the query set, so that $\phi_q$ is different from $\phi_s$ (i.e., large $D_{KL}(\phi_q \parallel \phi_s)$),
we compute the KL divergence between $\phi_c^i$ of each candidate video sample and $\phi_s$
(i.e., $D_{KL}(\phi_c^i \parallel \phi_s)$) that can be computed efficiently. 
(4) Finally, we can select the set of video samples that have the largest KL divergences w.r.t. $\phi_s$, to construct the query set.

In a similar manner, we can apply the above strategy to automatically  construct other query sets.
Note that different query sets can share common data samples.
Moreover, to help cover the wide range of possible conditional biases in the dataset, instead of fixing the support set and query sets during the whole training process, at the beginning of each training epoch, we \textit{randomly} select a part of the training set to re-construct the support set, and use the remaining part to automatically re-construct various query sets via the above strategy.
In this way, by performing \textit{meta training and testing}, during the whole training process, our model can learn to effectively handle various types of possible conditional biases.

\subsection{Training and Testing}
We can flexibly apply our framework to train the off-the-shelf VidSGG models.
During training, at each epoch, we first split the training set to construct a support set and a group of query sets as discussed above.
Then we perform \textit{meta training and testing} over the support set and query sets, to iteratively optimize the VidSGG model.
During testing, we can evaluate the trained model on the testing set in the conventional manner.

\section{Experiments}
We evaluate our framework on two datasets for two evaluation settings in VidSGG respectively:
ImageNet-VidVRD \cite{shang2017video} for video-level VidSGG, and
Action Genome \cite{ji2020action} for frame-level VidSGG.
More experiment results are in supplementary.

\textbf{ImageNet-VidVRD (VidVRD).} 
VidVRD dataset \cite{shang2017video} contains 1000 video samples with 35 object categories and 132 predicate categories.
For each video in VidVRD dataset, the model needs to predict a set of relation instances, and each relation instance contains a relation triplet with the subject and object trajectories.
Following \cite{shang2017video,shang2021video}, we use two evaluation protocols on this dataset: 
relation detection and relation tagging.
For relation detection,
we count a predicted relation instance as a correct one, if its relation triplet is the same with a ground truth, and their trajectory vIoU (volume IoU) of the subject and object are both larger than the threshold of 0.5.
In the same way as \cite{shang2017video,shang2021video}, 
we adopt Mean Average Precision (mAP), Recall@50 (R@50) and Recall@100 (R@100) to evaluate the model performance on relation detection.
While in relation tagging, for a predicted relation instance,  following \cite{shang2017video,shang2021video} we only consider the correctness of its relation triplet, and ignore the precision of its subject and object trajectories. 
The evaluation metrics of Precision@1 (P@1), Precision@5 (P@5) and Precision@10 (P@10) are used in relation tagging \cite{shang2017video,shang2021video}.

\textbf{Action Genome (AG).} 
AG dataset \cite{ji2020action} provides scene graph annotation for each video frame, i.e., the model needs to predict the scene graph of each frame.
AG dataset contains 234253 video frames with 35 object categories and 25 predicate categories.
Following \cite{teng2021target,cong2021spatial,ji2020action}, we evaluate models on three standard sub-tasks on this dataset:
predicate classification (PredCls), scene graph classification (SGCls) and scene graph detection (SGDet).
For these three sub-tasks, in line with \cite{teng2021target}, we use Recall (R@20, R@50), Mean Recall (MR@20, MR@50), m$\text{AP}_{rel}$ and wm$\text{AP}_{rel}$ to measure model performance.

\subsection{Implementation Details}
We conduct our experiments on an RTX 3090 GPU.
For experiments of video-level VidSGG on VidVRD dataset, we use the VidVRD-\uppercase\expandafter{\romannumeral2} network \cite{shang2021video} as the backbone of our framework, which exploits the spatio-temporal contexts via iterative relation inference.
For experiments of frame-level VidSGG on AG dataset, we use TRACE network \cite{teng2021target} as the backbone of our framework, which adaptively aggregates contextual information to infer the scene graph.

On these two datasets, at each training epoch, we randomly select $60\%$ of the training samples as the support set, and the remaining training samples are used to construct the query sets.
We set the size of each query set to 100 on VidVRD, and 200 on AG.
Note that in AG dataset, since the subject of all relation triplets is fixed to ``person", we skip the query sets for handling the conditional biases w.r.t. the prediction of subject 
(e.g., \textit{subject-centered} group) 
in this dataset.
We set the learning rate ($\alpha$) for \textit{meta training} to 0.0005, and the learning rate ($\beta$) for meta-optimization to 0.01. 

\subsection{Experimental Results}
On VidVRD dataset, compared to existing approaches, our method achieves the best performance across all metrics on both relation detection and relation tagging as shown in Table \ref{Tab:vidvrd}. 
This demonstrates that by reducing various types of conditional biases, our method can effectively enhance the model performance.
Moreover, we also compare our method to other debiasing methods in SGG, including two representative methods (Reweight \cite{tang2020unbiased} and TDE \cite{tang2020unbiased}) and a recently proposed one (DLFE \cite{chiou2021recovering}).
For Reweight, we follow the idea in \cite{tang2020unbiased}.
These methods use the same backbone (VidVRD-II \cite{shang2021video}) with ours.
The results in Table \ref{Tab:vidvrd} show that compared to these methods, our method achieves superior performance, demonstrating that by considering the spatio-temporal structure of VidSGG, our method can better handle the biases in this task.

On AG dataset, as shown in Table \ref{Tab:ag_recall} and Table \ref{Tab:ag_mean_recall_ap}, our method outperforms other methods on all metrics. 
Our method also outperforms other debiasing methods \cite{tang2020unbiased,chiou2021recovering} that use the same backbone (TRACE \cite{teng2021target}) with ours.
Moreover, note that the metric of Mean Recall is designed to measure the model performance considering the imbalanced data distribution \cite{tang2020unbiased,teng2021target}, and our method achieves more performance improvements on this metric, demonstrating that our framework can effectively mitigate the spatio-temporal conditional bias problem caused by biased data distribution in the dataset.

\begin{table}[t]
\caption{Comparison with state-of-the-arts on VidVRD dataset.}
\centering
\resizebox{0.5\textwidth}{!}{
\small
\setlength{\tabcolsep}{2pt}
\begin{tabular}{cccccccccccccc} \hline
Method &  \multicolumn{3}{c}{Relation Detection} & \multicolumn{3}{c}{Relation Tagging} \\ \cmidrule(lr){2-4} \cmidrule(lr){5-7}
& $\mathrm{mAP}$ & R@50 & R@100 & P@1 & P@5 & P@10\\ \hline\hline
VidVRD \cite{shang2017video}   & 8.58 & 5.54 & 6.37 & 43.00 & 28.90 & 20.80 \\
GSTEG \cite{tsai2019video} & 9.52 & 7.05 & 8.67 & 51.50 & 39.50 & 28.23 \\
VRD-GCN \cite{qian2019video} & 14.23 & 7.43 & 8.75 & 59.50 & 40.50 & 27.85 \\
VRD-GCN+siamese \cite{qian2019video} & 16.26 & 8.07 & 9.33 & 57.50 & 41.00 & 28.50 \\
VRD-STGC \cite{liu2020beyond}& 18.38 & 11.21 & 13.69 & 60.00 & 43.10 & 32.24 \\
VidVRD+MHA \cite{su2020video} &15.71 &7.40 &8.58 &40.00 &26.70 &18.25\\
VRD-GCN+MHA \cite{su2020video} &19.03 &9.53 &10.38 &57.50 &41.40 &29.45\\
TRACE \cite{teng2021target}& 17.57 & 9.08 & 11.15 & 61.00 & 45.30 & 33.50\\
Social Fabric \cite{chen2021social}& 20.08 & 13.73 & 16.88 & 62.50 & 49.20 & 38.45\\
IVRD \cite{li2021interventional} & 22.97 & 12.40 & 14.46 & 68.83 & 49.87 & 35.57\\ 
VidVRD-II \cite{shang2021video}& 29.37 & 19.63 & 22.92 & 70.40 & 53.88 & 40.16\\ 
VidVRD-II \cite{shang2021video} + Reweight \cite{tang2020unbiased} & 29.52 & 19.80 & 22.96 & 71.50 & 54.30 & 40.20 \\
VidVRD-II \cite{shang2021video} + TDE \cite{tang2020unbiased} 
& 29.78 & 19.90 & 23.04 & 72.50 & 54.50 & 40.65\\
VidVRD-II \cite{shang2021video} + DLFE \cite{chiou2021recovering} & 29.92 & 19.98 & 23.16 & 73.50 & 54.90 & 41.10\\ \hline
\textbf{Ours}  & \textbf{31.57} & \textbf{21.16} & \textbf{24.57} & \textbf{79.00} & \textbf{57.60} &\textbf{43.20} \\ \hline
\end{tabular}}
\label{Tab:vidvrd}
\end{table}

\begin{table}[t]
\parbox{.49\linewidth}{
\caption{We apply our framework on various models, and obtain consistent performance improvement on VidVRD dataset.}
\centering
\resizebox{0.49\textwidth}{!}
{\small
\setlength{\tabcolsep}{2pt}
\begin{tabular}{cccccccccccccc}
\hline
Method &  \multicolumn{3}{c}{Relation Detection} & \multicolumn{3}{c}{Relation Tagging} \\ \cmidrule(lr){2-4} \cmidrule(lr){5-7}
& $\mathrm{mAP}$ & R@50 & R@100 & P@1 & P@5 & P@10\\
 \hline\hline
VRD-STGC\cite{liu2020beyond}& 18.38 & 11.21 & 13.69 & 60.00 & 43.10 & 32.24 \\
VRD-STGC + Ours &20.76 &12.62 &15.78 &65.50 &44.90 &33.15\\
\hline
Independent baseline\cite{shang2021video}& 27.49 & 18.18 & 21.28 & 67.10 & 50.18 & 38.02\\
Independent baseline + Ours & 30.02 & 19.86 & 23.10 & 75.50 & 53.60 & 40.80 \\
\hline
VidVRD-II\cite{shang2021video}& 29.37 & 19.63 & 22.92 & 70.40 & 53.88 & 40.16\\
VidVRD-II + Ours & 31.57 & 21.16 & 24.57 & 79.00 & 57.60 & 43.20 \\
\hline
\end{tabular}}
\label{Tab:ablation_study_2}
}
\hspace{0.01\linewidth}
\parbox{0.5\linewidth}{
\caption{We apply our framework on different SOTA models for image SGG, and obtain consistent performance improvement.}
\centering
\resizebox{0.5\textwidth}{!}
{
\small
\setlength{\tabcolsep}{2pt}
\begin{tabular}{ccccccccccccccccc}
\hline
Method &  \multicolumn{3}{c}{SGGen} & \multicolumn{3}{c}{SGCls} & \multicolumn{3}{c}{PredCls}\\ \cmidrule(lr){2-4} \cmidrule(lr){5-7} \cmidrule(lr){8-10}
& mR@20 & mR@50 & mR@100 & mR@20 & mR@50 & mR@100 & mR@20 & mR@50 & mR@100\\
\hline\hline
VCTree \cite{tang2019learning} &5.2&7.1&8.3&9.1&11.3&12.0& 14.1&17.7&19.1 \\
VCTree+Ours & 10.1 & 13.1 & 15.4 & 16.9 & 19.6 & 20.6 & 25.7 & 29.8 & 31.4\\ \hline 
BGNN \cite{li2021bipartite}  & - & 10.7 & 12.6 & - & 14.3 & 16.5 & - & 30.4 & 32.9\\
BGNN + Ours  & 11.1 & 14.2 & 16.4 & 15.9 & 17.4 & 18.6 & 27.3 & 31.6 & 34.1\\
\hline
\end{tabular}}
\label{Tab:ablation_image}
}
\end{table}

\begin{table}[t]
\caption{Recall (\%) of various models on AG dataset following the setting in \cite{teng2021target}.}
\centering
\resizebox{0.8\textwidth}{!}
{
\small
\setlength{\tabcolsep}{2pt}
\begin{tabular}{cccccccccccccc}
\hline
\begin{tabular}[c]{@{}c@{}}Top k Predictions \\ for Each Pair\end{tabular} & Method &  \multicolumn{4}{c}{SGDet} & \multicolumn{4}{c}{SGCls} & \multicolumn{4}{c}{PredCls} \\ \cmidrule(lr){3-6} \cmidrule(lr){7-10} \cmidrule(lr){11-14}
& & \multicolumn{2}{c}{image}  & \multicolumn{2}{c}{video} & \multicolumn{2}{c}{image} & \multicolumn{2}{c}{video} & \multicolumn{2}{c}{image} & \multicolumn{2}{c}{video} \\ \cmidrule(lr){3-4} \cmidrule(lr){5-6} \cmidrule(lr){7-8} \cmidrule(lr){9-10} \cmidrule(lr){11-12} \cmidrule(lr){13-14}
 & & R@20   & R@50 & R@20 & R@50 & R@20 & R@50 & R@20 & R@50 & R@20 & R@50 & R@20 & R@50 \\
 \hline\hline
k=7 & Freq Prior  \cite{zellers2018neural}   
& 34.41 & 44.34 & 32.50 & 41.11
& 45.10 & 48.87 & 44.47 & 46.39
& 87.95 & 93.02 & 86.01 & 88.59\\
       & G-RCNN  \cite{yang2018graph}  
& 34.28 & 44.47 & 32.60 & 41.29 
& 45.57 & 49.75 & 45.11 & 47.22
& 88.73 & 93.73 & 86.28 & 88.93\\
       & RelDN  \cite{zhang2019graphical}      
& 34.92 & 45.27 & 33.18 & 42.10 
& 46.47 & 50.31 & 45.87 & 47.78
& 90.89 & 96.09 & 88.77 & 91.43 \\
       & TRACE  \cite{teng2021target} 
& 35.09 & 45.34 & 33.38 & 42.18 
& 46.66 & 50.46 & 46.03 & 47.92
& 91.60 & 96.35 & 89.31 & 91.72 \\  
       & TRACE \cite{teng2021target} + Reweight \cite{tang2020unbiased}
& 35.15 & 45.37 & 33.42 & 42.24 
& 46.68 & 50.50 & 46.07 & 47.94
& 91.61 & 96.35 & 89.32 & 91.74  \\
       & TRACE \cite{teng2021target} + TDE \cite{tang2020unbiased}
& 35.20 & 45.41 & 33.49 & 42.30 
& 46.71 & 50.55 & 46.12 & 48.00
& 91.63 & 96.36 & 89.32 & 91.76  \\
       & TRACE \cite{teng2021target} + DLFE \cite{chiou2021recovering}
& 35.29 & 45.47 & 33.58 & 42.41 
& 46.75 & 50.63 & 46.18 & 48.04
& 91.64 & 96.36 & 89.35 & 91.77  \\
       \cline{2-14}
       & \textbf{Ours} 
& \textbf{36.59} & \textbf{47.00} & \textbf{34.88} & \textbf{43.81} 
& \textbf{47.40} & \textbf{51.06} & \textbf{46.71} & \textbf{48.56}
& \textbf{91.74} & \textbf{96.43} & \textbf{89.44} & \textbf{91.85} \\
 \hline
k=6 & Freq Prior  \cite{zellers2018neural}   
& 34.47 & 43.69 & 32.38 & 40.24 
& 44.90 & 47.15 & 43.57 & 44.63
& 85.89 & 89.43 & 83.33 & 84.99 \\
       & G-RCNN \cite{yang2018graph}   
& 34.60 & 43.98 & 32.75 & 40.65 
& 45.82 & 48.31 & 44.60 & 45.77 
& 87.03 & 90.60 & 84.02 & 85.74 \\
       & RelDN  \cite{zhang2019graphical}      
& 35.22 & 44.94 & 33.39 & 41.64 
& 46.76 & 49.11 & 45.48 & 46.57
& 89.63 & 93.56 & 87.01 & 88.86 \\
       & TRACE  \cite{teng2021target} 
& 35.41 & 45.06 & 33.59 & 41.76 
& 47.00 & 49.32 & 45.71 & 46.79 
& 90.34 & 93.94 & 87.56 & 89.24 \\ 
       & TRACE \cite{teng2021target} + Reweight \cite{tang2020unbiased}
& 35.44 & 45.10 & 33.64 & 41.83 
& 47.01 & 49.35 & 45.73 & 46.82
& 90.36 & 93.95 & 87.58 & 89.27  \\
       & TRACE \cite{teng2021target} + TDE \cite{tang2020unbiased}
& 35.49 & 45.16 & 33.68 & 41.90 
& 47.04 & 49.36 & 45.76 & 46.89
& 90.37 & 93.96 & 87.61 & 89.27  \\
       & TRACE \cite{teng2021target} + DLFE \cite{chiou2021recovering}
& 35.56 & 45.28 & 33.76 & 41.99 
& 47.08 & 49.41 & 45.83 & 46.92
& 90.39 & 93.99 & 87.65 & 89.29  \\       
       \cline{2-14}
       & \textbf{Ours}  
& \textbf{36.80} & \textbf{46.73} & \textbf{34.99} & \textbf{43.39} 
& \textbf{47.66} & \textbf{49.96} & \textbf{46.41} & \textbf{47.47} 
& \textbf{90.49} & \textbf{94.11} & \textbf{87.78} & \textbf{89.50} \\
 \hline
\end{tabular}}
\label{Tab:ag_recall}
\end{table}

\begin{table}[t]
\caption{Mean Recall (\%) and Average Precision (\%) of various models on AG dataset following the setting in \cite{teng2021target}.}
\centering
\resizebox{0.8\textwidth}{!}{
\small
\setlength{\tabcolsep}{2pt}
\begin{tabular}{cccccccccccccc}
\hline
Method &  \multicolumn{4}{c}{SGDet} & \multicolumn{4}{c}{SGCls} & \multicolumn{4}{c}{PredCls} \\ \cmidrule(lr){2-5} \cmidrule(lr){6-9} \cmidrule(lr){10-13}
& \multicolumn{2}{c}{Mean Recall}  & \multicolumn{2}{c}{Average Precision} & \multicolumn{2}{c}{Mean Recall} & \multicolumn{2}{c}{Average Precision} & \multicolumn{2}{c}{Mean Recall} & \multicolumn{2}{c}{Average Precision} \\ \cmidrule(lr){2-3} \cmidrule(lr){4-5} \cmidrule(lr){6-7} \cmidrule(lr){8-9} \cmidrule(lr){10-11} \cmidrule(lr){12-13}
& @20 & @50 & $\mathrm{mAP}_{r}$ & $\mathrm{wmAP}_{r}$ & @20 & @50 & $\mathrm{mAP}_{r}$ & $\mathrm{wmAP}_{r}$ & @20 & @50 & $\mathrm{mAP}_{r}$ & $\mathrm{wmAP}_{r}$ \\
 \hline\hline
Freq Prior
\cite{zellers2018neural}   
& 24.89 & 34.07 & 9.45 & 15.58 
& 34.30 & 36.96 & 14.29 & 22.68
& 55.17 & 63.67 & 33.10 & 65.92 \\
        G-RCNN
        \cite{yang2018graph}  
& 27.79 & 34.99 & 11.76 & 15.90 
& 36.19 & 38.29 & 17.64 & 22.53 
& 56.32 & 61.31 & 41.21 & 70.89 \\
        RelDN
        \cite{zhang2019graphical}      
& 30.39 & 39.53 & 12.93 & 15.94 
& 39.92 & 41.93 & 20.07 & 23.88 
& 59.81 & 63.47 & 50.08 & 72.26 \\
        TRACE
        \cite{teng2021target} 
& 30.84 & 40.12 & 13.43 & 16.56 
& 41.19 & 43.21 & 20.71 & 24.61
& 61.80 & 65.37 & 53.27 & 75.45 \\ 
      TRACE \cite{teng2021target} + Reweight \cite{tang2020unbiased}
& 30.87 & 40.21 & 13.44 & 16.59 
& 41.31 & 43.44 & 20.75 & 24.63
& 61.97 & 65.77 & 53.30 & 75.46 \\
      TRACE \cite{teng2021target} + TDE \cite{tang2020unbiased}
& 31.01 & 40.40 & 13.47 & 16.60 
& 41.56 & 43.70 & 20.79 & 24.66
& 62.12 & 65.89 & 53.34 & 75.50 \\
      TRACE \cite{teng2021target} + DLFE \cite{chiou2021recovering}
& 31.24 & 40.75 & 13.48 & 16.62 
& 41.77 & 43.98 & 20.83 & 24.70
& 62.44 & 66.31 & 53.35 & 75.52 \\
      \hline
        \textbf{Ours}  
& \textbf{32.43} & \textbf{43.13} & \textbf{14.00} & \textbf{17.47} 
& \textbf{43.43} & \textbf{47.26} & \textbf{21.25} & \textbf{25.32} 
& \textbf{67.67} & \textbf{75.72} & \textbf{53.88} & \textbf{75.96} \\
 \hline
\end{tabular}}
\label{Tab:ag_mean_recall_ap}
\end{table}

\subsection{Ablation Studies}
We conduct extensive ablation experiments to evaluate our framework on VidVRD dataset.

\textbf{Impact of different backbone networks.} 
To validate the general effectiveness of our framework, 
we apply it on different models \cite{liu2020beyond,shang2021video}, 
and obtain consistent performance improvement as shown in Table \ref{Tab:ablation_study_2},
showing our framework can be flexibly applied on various models to improve their performance.

\textbf{Impact of spatio-temporal conditional biases.}
To investigate the impact of spatial and temporal conditional biases on model performance,
we evaluate the following variants.
For spatial conditional biases, we test 4 variants.

\begin{wraptable}[11]{r}{0.6\linewidth}
\caption{We evaluate various variants to investigate the impact of each group of spatio-temporal conditional biases.}
\centering
\resizebox{0.5\textwidth}{!}
{
\small
\setlength{\tabcolsep}{2pt}
\begin{tabular}{lcccccccccccc} \hline
Method &  \multicolumn{3}{c}{Relation Detection} & \multicolumn{3}{c}{Relation Tagging} \\ \cmidrule(lr){2-4} \cmidrule(lr){5-7}
& $\mathrm{mAP}$ & R@50 & R@100 & P@1 & P@5 & P@10\\ \hline\hline
Baseline (VidVRD-II) & 29.37  &19.63  &22.92  &70.40  &53.88  &40.16 \\ \hline
w/o Spatial Level (all)  &30.49 &20.35 &23.61 &75.50 &55.30 &41.50\\ 
w/o Predicate-centered  &30.98 &20.67 &23.94 &76.50 &56.40 &42.35\\
w/o Subject-centered & 31.10 &20.87 &24.04 &78.00 &56.80 &42.75 \\ 
w/o Object-centered &31.05 &20.80 &23.98 &77.50 &56.60 &42.60 \\ \hline
w/o Temporal Level (all) &30.47  &20.48  &23.63  &75.00  &55.60  &41.70 \\
w/o Forward Case & 30.94  &20.78 &23.95  &76.50  &56.50  &42.60 \\
w/o Backward Case &31.00  &20.75  &24.01  &77.50  &56.80  &42.45 \\ \hline
\textbf{Ours}  & \textbf{31.57} & \textbf{21.16} & \textbf{24.57} & \textbf{79.00} & \textbf{57.60} &\textbf{43.20} \\ \hline
\end{tabular}}
\label{Tab:ablation_study_1}
\end{wraptable}

Specifically, one model variant (\textit{w/o Spatial Level (all)}) ignores \textit{all} groups of spatial conditional biases and handles only temporal conditional biases, i.e., optimizing the model without the query sets for handling \textit{all} types of spatial conditional biases.
Moreover, as discussed in \ref{sec:dataset_split}, we have 3 groups of spatial conditional biases, i.e., 
the conditional bias between \textit{predicate}/\textit{subject}/\textit{object} and their corresponding spatial contexts.
Thus to explore the impact of each group of conditional biases, we correspondingly implement 3 variants (\textit{w/o Predicate-centered}, \textit{w/o Subject-centered} and \textit{w/o Object-centered}), and each variant ignores the corresponding group of query sets during the model training.

Similarly, for temporal conditional biases, we have 3 variants.
One model variant (\textit{w/o Temporal Level (all)}) ignores \textit{all} groups of temporal conditional biases, and handles only spatial conditional biases.
Furthermore, since we have 2 groups of temporal conditional biases, i.e., the conditional bias between the current triplet and the triplets appear before or appear after,
we evaluate 2 more variants (\textit{w/o Forward Case} and \textit{w/o Backward Case}).

As shown in Table \ref{Tab:ablation_study_1}, ignoring any group of conditional biases would lead to performance drop compared to our framework, showing that each group of conditional biases can affect the model performance. 
More ablation study and qualitative results are in our supplementary.

\begin{wraptable}[8]{r}{0.6\linewidth}
\caption{
Experiment results of ours and other debiasing methods in image SGG.}
\centering
\resizebox{0.6\textwidth}{!}
{
\small
\setlength{\tabcolsep}{2pt}
\begin{tabular}{cccccccccccccc}
\hline
Method &  \multicolumn{2}{c}{SGGen} & \multicolumn{2}{c}{SGCls} & \multicolumn{2}{c}{PredCls}\\ \cmidrule(lr){2-3} \cmidrule(lr){4-5} \cmidrule(lr){6-7}
& mR@20 & mR@50 & mR@20 & mR@50 & mR@20 & mR@50\\ \hline\hline
VCTree \cite{tang2019learning} & 5.2 & 7.1 & 9.1 & 11.3 & 14.1 & 17.7 \\
VCTree+Reweight \cite{tang2020unbiased} & 6.6 & 8.7 & 10.6 & 12.5 & 16.3 & 19.4\\
VCTree+TDE \cite{tang2020unbiased} & 6.8 & 9.5 & 11.2 & 15.2 & 19.2 & 26.2\\
VCTree+DLFE \cite{chiou2021recovering} & 8.6 & 11.8 & 15.8 & 18.9 & 20.8 & 25.3\\ \hline
VCTree+Ours & 13.1 & 15.4 & 19.6 & 20.6 & 29.8 & 31.4\\ \hline 
\end{tabular}}
\label{Tab:ablation_image_2}
\end{wraptable}

\subsection{Experiments on Image SGG}
Besides the VidSGG task, there can also exist spatial conditional biases in the task of image SGG.
Thus if we remove the query sets for handling the temporal conditional biases, our framework can then be adapted to handle the image SGG task.
Therefore, we also evaluate our method on the widely used SGG benchmark: Visual Genome \cite{krishna2017visual}, by constructing and incorporating only the query sets for handling the spatial conditional biases.
As shown in Table \ref{Tab:ablation_image},  we apply our framework on different SGG models \cite{li2021bipartite,tang2019learning}, and consistently enhance their performances. 
Besides, as shown in Table \ref{Tab:ablation_image_2}, based on the same backbone (VCTree \cite{tang2019learning}), our method achieves better performance than other debiasing strategies.

\section{Conclusion}
To address the spatio-temporal conditional bias problem in VidSGG, we propose a novel Meta Video Scene Graph Generation (\textbf{MVSGG}) framework.
By constructing a support set and various query sets w.r.t. various types of conditional biases, and optimizing the model on these constructed sets via \textit{meta training and testing},
our framework can effectively train the model to handle various types of conditional biases.
Moreover, our framework is general, and can be flexibly applied to various models. 
Our framework achieves superior performance.

\section*{Acknowledgement}
This work is supported by National Research Foundation, Singapore under its AI Singapore Programme (AISG Award No: AISG-100E-2020-065), Ministry of Education Tier 1 Grant and SUTD Startup Research Grant.
\clearpage
\bibliographystyle{splncs04}
\bibliography{main}
\end{document}